\documentclass[11pt,a4paper]{article}
\usepackage[hyperref]{naaclhlt2018}
\usepackage{times}
\usepackage{latexsym}
\usepackage{amsmath}
\usepackage{url}
\usepackage{graphicx}
\usepackage[font=small,skip=0pt]{caption}
\usepackage{amsfonts}
\usepackage{bbm}
\usepackage{multirow}
\usepackage{adjustbox}
\DeclareMathOperator*{\argmax}{argmax}

\aclfinalcopy % Uncomment this line for the final submission
\title{Sentence Simplification with Memory-Augmented Neural Networks}
\author{Tu Vu$^{1}$, Baotian Hu$^{2}$, Tsendsuren Munkhdalai$^{3}$ and Hong Yu$^{1,2}$ \\
  $^{1}$University of Massachusetts Amherst, Amherst, MA 01003, USA\\
  \tt tuvu@cs.umass.edu\\
  $^{2}$University of Massachusetts Medical School, Worcester, MA 01655, USA\\
  \tt \{baotian.hu,hong.yu\}@umassmed.edu\\
  $^{3}$Microsoft Research, Montr\'eal, QC H3A 3H3, Canada\\
  \tt tsendsuren.munkhdalai@microsoft.com\\ \\}

\date{}

\begin{document}
\maketitle
\begin{abstract}
  Sentence simplification aims to simplify the content and structure of complex sentences, and thus make them easier to interpret for human readers, and easier to process for downstream NLP applications. %Most successful methods for the task have relied on traditional machine translation models. However, 
  Recent advances in neural machine translation have paved the way for novel approaches to the task. In this paper, we adapt an architecture with augmented memory capacities called Neural Semantic Encoders \cite{Munkhdalai:17a} for sentence simplification. Our experiments demonstrate the effectiveness of our approach on different simplification datasets, both in terms of automatic evaluation measures and human judgments.
   %several neural sequence to sequence models adapted for sentence simplification utilizing different neural network architectures for the encoder and decoder, such as a memory-augmented recurrent neural network, or a tree-structured recursive neural network. Our experiments demonstrate the superiority of our approaches over the state-of-the-art simplification systems on different datasets, both in terms of automatic evaluation measures and human judgments.
\end{abstract}

\section{Introduction}
%The need to facilitate access to reading materials for people who have difficulties reading and understanding a text is obvious. According to the American Dyslexia Association, over 40 million adults and at least 20\% of school-aged children in the United States are suffering from dyslexia. 
The goal of sentence simplification is to compose complex sentences into simpler ones so that they are more comprehensible and accessible, while still retaining the original information content and meaning. %For instance, the sentence ``\textit{President Trump has a plethora of movie and television credits that most budding actors can only dream of.}" can be split into two simpler ones ``\textit{President Trump has been in a lot of movies and TV shows. Most young actors can only dream of this.}" 
%In practice, simplification is implemented using several major operations: \textit{splitting} a long and complex sentence into several simpler ones; \textit{deletion} of unimportant parts of a sentence; \textit{reordering} parts in a sentence or the split sentences; \textit{substitution} of difficult words/phrases with more common synonyms ~\cite{Zhu:10}. Xu et al. ~\shortcite{Xu:16} categorize reordering, lexical substitutions and syntactic transformations into \textit{paraphrasing}.
Sentence simplification has a number of practical applications.  On one hand, it provides reading aids for people with limited language proficiency%, including children, low-literacy readers, and non-native speakers 
~\cite{Watanabe:09,Siddharthan:03}, or for patients with linguistic and cognitive disabilities ~\cite{Carroll:99}. On the other hand, it can improve the performance of other NLP tasks ~\cite{Chandrasekar:96, Knight:00,Klebanov:04}.%, or information extraction ~\cite{Niklaus:16}.%, semantic role labeling ~\cite{Vickrey:08}, or sentence fusion ~\cite{Filippova:08}.

Prior work has explored monolingual machine translation (MT) approaches, 
% where the output should be equivalent in meaning to the input, but simpler in terms of vocabulary and sentence structure.  The key idea of these approaches is to utilize corpora of simplified texts,
utilizing corpora of simplified texts, e.g., Simple English Wikipedia (SEW), and making use of statistical MT models, such as phrase-based MT (PBMT) ~\cite{Stajner:15,Coster:11,Wubben:12}, tree-based MT (TBMT) ~\cite{Zhu:10,Woodsend:11}, or syntax-based MT (SBMT) \cite{Xu:16}.

Inspired by the success of neural MT ~\cite{Sutskever:14,Cho:14}, recent work has started exploring neural simplification with sequence to sequence (Seq2seq) models, also referred to as encoder-decoder models.  Nisioi et al. ~\shortcite{Nisioi:17} implemented a standard LSTM-based Seq2seq model %~\cite{Luong:15}
and found that they outperform PBMT, SBMT, and unsupervised lexical simplification approaches. Zhang and Lapata ~\cite{Zhang:17} viewed the encoder-decoder model as an agent and employed a deep reinforcement learning framework in which the reward has three components capturing key aspects of the target output: simplicity, relevance, and fluency. %They also examined a linear combination of the model and a lexical simplification model. 
 %implemented by recurrent neural networks (RNNs), either alone \cite{Nisioi:17} or coupled with a deep reinforcement learning framework \cite{Zhang:17}.
 
The common practice for Seq2seq models is to use recurrent neural networks (RNNs) with Long Short-Term Memory \cite[LSTM,][]{Hochreiter:97} or Gated Recurrent Unit \cite[GRU,][]{Cho:14} for the encoder and decoder \cite{Nisioi:17,Zhang:17}. These architectures were designed to %solve the vanishing and exploding gradient problems, and are 
be capable of memorizing long-term dependencies across sequences. Nevertheless, their memory is typically small and might not be enough for the simplification task, where one is confronted with long and complicated sentences. %Furthermore, they lacks the capacity to capture different aspects of compositionality in language that may be useful for simplification.

In this study, we go beyond the conventional LSTM/GRU-based Seq2seq models and propose to use a memory-augmented RNN architecture called Neural Semantic Encoders (NSE). This architecture has been shown to be effective in a wide range of NLP tasks \cite{Munkhdalai:17a}. The contribution of this paper is twofold:
\begin{figure}[h!]
\centering
\includegraphics[scale=0.28]{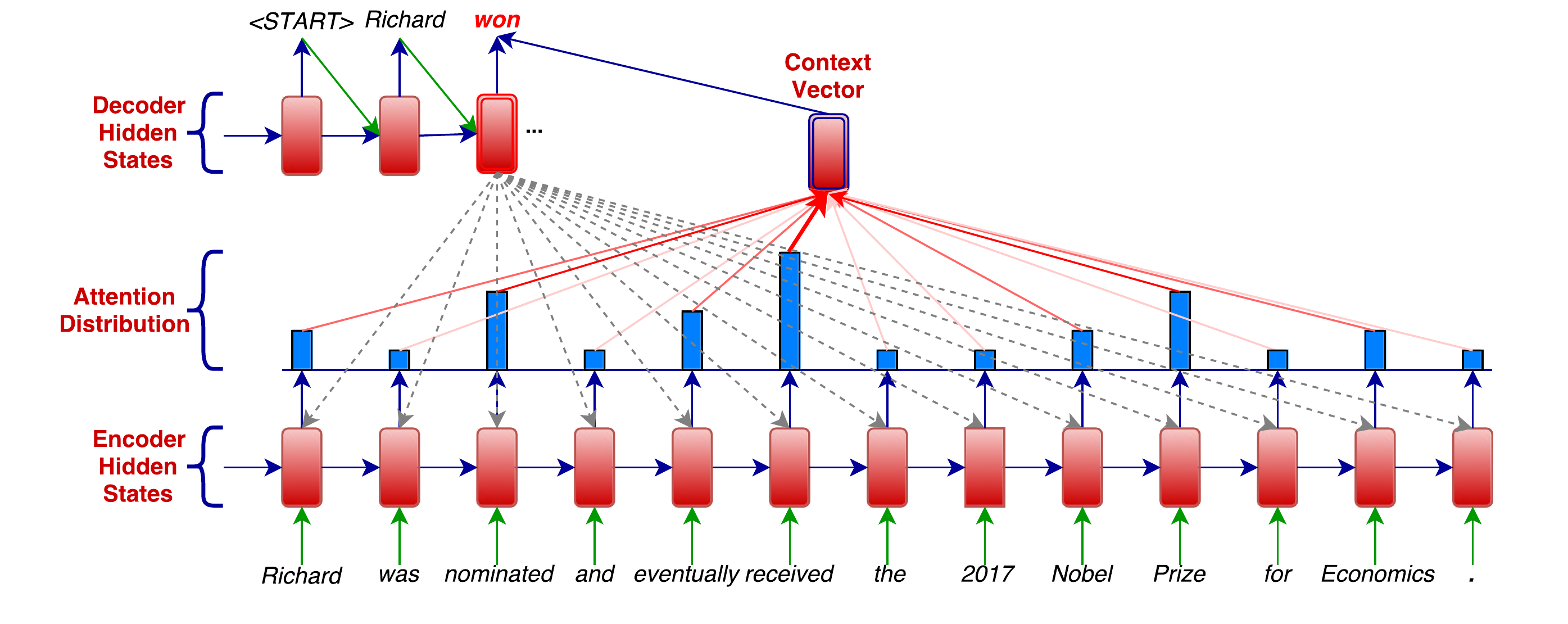}
\caption{Attention-based encoder-decoder model. The model may attend to relevant positions in the source sentence while decoding the simplification, e.g., to generate the target word \textit{won} the model may attend to the words \textit{received}, \textit{nominated} and \textit{Prize} in the source sentence.}
\label{fig1}
\vspace*{-3mm}
\end{figure}

(1) First, we present a novel simplification model which is, to the best of our knowledge, the first model that use memory-augmented RNN for the task. We investigate the effectiveness of neural Seq2seq models when different neural architectures for the encoder are considered. Our experiments reveal that the \textsc{NseLstm} model that uses an NSE as the encoder and an LSTM as the decoder performed the best among these models, improving over strong simplification systems. (2) Second, we perform an extensive evaluation of various approaches proposed in the literature on different datasets. Results of both automatic and human evaluation show that our approach is remarkably effective for the task, significantly reducing the reading difficulty of the input, while preserving grammaticality and the original meaning. We further discuss some advantages and disadvantages of these approaches.

\section{Neural Sequence to Sequence Models}
%Unlike the traditional MT-based simplification systems which often consist of several components that are tuned separately, neural simplification aims at building a single end-to-end trainable model that reads an input complex sentence and outputs its simpler version. 
%We first present a basic encoder-decoder framework for the task and then describe our approach.

\subsection{Attention-based Encoder-Decoder Model}
\label{sec3.1}
Our approach is based on an attention-based Seq2seq model \cite{Bahdanau:15} %which learns to align and simplify jointly, and is 
 (Figure \ref{fig1}). Given a complex source sentence $\mathcal{X} = x_{1:T_x}$, the model learns to generate its simplified version $\mathcal{Y} = y_{1:T_y}$. The encoder reads through $\mathcal{X}$ and computes a sequence of hidden states $h_{1:T_x}$:
\begin{center}
$h_t = \mathcal{F}^{enc}(h_{t-1}, x_t)$,
\end{center}
where $\mathcal{F}^{enc}$ is a non-linear activation function (e.g., LSTM), $h_t$ is the hidden state at time $t$. Each time the model generates a target word $y_t$, the decoder looks at a set of positions in the source sentence where the most relevant information is located. %This is desirable for simplification as it allows words to be simplified based on the context they are used in. 
Specifically, another non-linear activation function $F^{dec}$ is used for the decoder where the hidden state $s_t$ at time $t$ is computed by:
\begin{center}
$s_t = \mathcal{F}^{dec}(s_{t-1}, y_{t-1}, c_t)$.
\end{center}
Here, the context vector $c_t$ is computed as a weighted sum of the hidden vectors $h_{1:T_x}$:
\begin{center}
$c_t = \sum\limits_{i=1}^{T_x} \alpha_{ti}h_i$, \hspace*{8mm}
$\alpha_{ti} = \frac{\text{exp} (s_{t-1} \odot h_i)}{ \sum\limits_{j=1}^{T_x} \text{exp} (s_{t-1} \odot h_j)}$,
\end{center}
where $\odot$ is the dot product of two vectors. %Intuitively, the alignment score $\alpha_{ti}$ tells us ``how well the inputs around position $i$ and the output at position $t$ match'' \cite{Bahdanau:15}. 
Generation is conditioned on $c_t$ and all the previously generated target words $y_{1:t-1}$:
\begin{center}
$P(\mathcal{Y}|\mathcal{X}) = \prod\limits_{t=1}^{T_y} P(y_t | \{y_{1:t-1}\}, c_t)$,\\[1mm]
$P(y_t | \{y_{1:t-1}\}, c_t) =  \mathcal{G}(y_{t-1}, s_t, c_t)$, 
\end{center} 
where $\mathcal{G}$ is some non-linear function. %that returns the probability of $y_t$. 
The training objective is to minimize the cross-entropy loss of the training source-target pairs.
%Seq2seq models often generate many out-of-vocabulary tokens $\langle unk \rangle$, affecting the quality of simplifications. Assuming the output at position $t$ is an $\langle unk \rangle$ token, in a similar vein to \cite{Jean:15,Luong:15}, we replace it with the source word $x_k$ with the highest alignment score $\alpha_{ti}$, i.e., $k = \argmax \limits_{i} (\alpha_{ti})$.
\subsection{Neural Semantic Encoders}
An RNN allows us to compute a hidden state $h_t$ of each word summarizing the preceding words $x_{1:t}$, but not considering the following words $x_{t+1:T_x}$ that might also be useful for simplification. An alternative approach is to use a bidirectional-RNN \cite{Schuster:97}. %where the hidden state $h_t$ is the concatenation of the hidden states produced by a \textit{forward} RNN that reads $\mathcal{X}$ as it is ordered, and a \textit{backward} RNN that reads $\mathcal{X}$ in the reverse order, $h_t = \overrightarrow{h_t} \oplus \overleftarrow{h_t}$.
Here, we propose to use Neural Semantic Encoders \cite[NSE,][]{Munkhdalai:17a}. During each encoding time step $t$, we compute a memory matrix $M_t \in \mathbb{R}^{T_x \times D}$ where $D$ is the dimensionality of the word vectors. This matrix is initialized with the word vectors %(each row $i$ of $M_0$ corresponds to a word vector $x_i$) 
and is refined over time through NSE's %\textit{read}, \textit{compose}, and \textit{write} 
functions to gain a better understanding of the input sequence. Concretely, NSE sequentially reads the tokens $x_{1:T_x}$ with its \textit{read} function:
\begin{center}
$r_t = \mathcal{F^\textit{{enc}}_\textit{read}}(r_{t-1}, x_t)$, 
\end{center} 
where $\mathcal{F^\textit{{enc}}_\textit{read}}$ is an LSTM, $r_t \in \mathbb{R}^D$ is the hidden state at time $t$. Then, a \textit{compose} function is used to compose $r_t$ with relevant information retrieved from the memory at the previous time step, $M_{t-1}$:
\begin{center}
$c_t = \mathcal{F^\textit{{enc}}_\textit{compose}}(r_t, m_t)$, 
\end{center}
where $\mathcal{F^\textit{{enc}}_\textit{compose}}$ is a multi-layer perceptron with one hidden layer, $c_t \in \mathbb{R}^{2D}$ is the output vector, and $m_t \in \mathbb{R}^{D}$ is a linear combination of the memory slots of $M_{t-1}$, weighted by $\sigma_{ti} \in \mathbb{R}$:
\begin{center}
$m_t = \sum \limits_{i=1}^{T_x} \sigma_{ti}M_{t-1,i}$, \hspace*{5mm}
$\sigma_{ti} = \frac{\text{exp} (r_t \odot M_{t-1,i})}{ \sum\limits_{j=1}^{T_x} \text{exp} (r_t \odot M_{t-1,j})}.$
\end{center}
Here, $M_{t-1,i}$ is the $i^{th}$ row of the memory matrix at time $t-1$, $M_{t-1}$. Next, a \textit{write} function is used to map $c_t$ to the encoder output space:
\begin{center}
$w_t = \mathcal{F^\textit{{enc}}_\textit{write}}(w_{t-1}, c_t)$, 
\end{center}
where $\mathcal{F^\textit{{enc}}_\textit{write}}$ is an LSTM, $w_t \in \mathbb{R}^{D}$ is the hidden state at time $t$. Finally, the memory is updated accordingly. The retrieved memory content pointed by $\sigma_{ti}$ is erased and the new content is added:
%\[
%M_{t} =  (\mathbbm{1} - \sigma_{t} \otimes \textbf{e}_D) \circ %M_{t-1} + (\sigma_{t} \otimes \textbf{e}_D) \circ (w_t \otimes %\textbf{e}_{T_x})^T.\]
%Here, $\sigma_t = (\sigma_{t1}, \sigma_{t2},\ldots,\sigma_{tT_x})$, $\otimes$ is the outer product of two vectors, $\circ$ is the Hadamard product of two matrices,  $\mathbbm{1}$ is the matrix of ones of size $T_x \times D$, and $\textbf{e}_N$ is the vector of ones of size $N$. This equation can be rewritten in terms of $M_{t,i}$
%as:
\begin{center}
$M_{t,i} =  (1 - \sigma_{ti})M_{t-1,i} + \sigma_{ti}w_t.$
\end{center}
NSE gives us unrestricted access to the entire source sequence stored in the memory. As such, the encoder may attend to relevant words when encoding each word. The sequence $w_{1:T_x}$ is then used as the sequence $h_{1:T_x}$ in Section \ref{sec3.1}.

%%Another NSE can be used for the decoder:
%\begin{center}
%$s_t =  \mathbb{NSE}(s_{t-1}, y_{t-1}, M_t)$
%\end{center}
% In this case, the decoder NSE's memory is initialized with the encoder's memory. It accesses and updates this memory with its own \textit{read}, \textit{compose}, \textit{write} functions to generate the target sequence:
%\begin{center}
%$r_t = \mathcal{F^\textit{{dec}}_\textit{read}}(r_{t-1}, s_{t-1}, y_{t-1})$,\\[2mm]
%$c_t = \mathcal{F^\textit{{dec}}_\textit{compose}}(r_t, m_t)$, \\[2mm]
%$s_t = w_t = \mathcal{F^\textit{{dec}}_\textit{write}}(w_{t-1}, c_t)$.
%\end{center}
%Here, the computations are similar to those previously described.
\subsection{Decoding}
We differ from the approach of Zhang et al. \shortcite{Zhang:17} in the sense that we implement both a greedy strategy and a beam-search strategy to generate the target sentence. Whereas the greedy decoder always chooses the simplification candidate with the highest log-probability, the beam-search decoder keeps a fixed number (beam) of the highest scoring candidates at each time step. We report the best simplification among the outputs based on automatic evaluation measures.
%We pick the simplification with the highest log-probability out of the final candidate list. 

\section{Experimental Setup}
%In this section we present our experimental setup for evaluating the performance of our simplification systems. We give details on the simplification corpora and the models we used, and explain how the system outputs were evaluated.
\subsection{Datasets}
Following \cite{Zhang:17}, we experiment on three simplification datasets, namely: (1) \textit{Newsela}  \cite{Xu:15}, a high-quality simplification corpus of news articles composed by Newsela\footnote{https://newsela.com} professional editors for children at multiple grade levels. We used the split of the data in \cite{Zhang:17}, i.e., 94,208/1,129/1,077 pairs for train/dev/test. (2) \textit{WikiSmall} \cite{Zhu:10},  which contains aligned complex-simple sentence pairs from English Wikipedia (EW) and SEW. %and has become a benchmark for evaluating simplification systems. %\cite{Woodsend:11,Wubben:12,Narayan:14,Zhang:17}. 
The dataset has 88,837/205/100 pairs for train/dev/test. (3) \textit{WikiLarge} \cite{Zhang:17}, a larger corpus in which the training set is a mixture of three Wikipedia datasets in \cite{Zhu:10,Woodsend:11,Kauchak:13}%, containing aligned sentence pairs from EW-SEW or SEW revision histories
, and the development and test sests are complex sentences taken from \textit{WikiSmall}, each has 8 simplifications written by Amazon Mechanical Turk workers \cite{Xu:16}. The dataset has 296,402/2,000/359 pairs for train/dev/test. Table \ref{tbl1} provides statistics on the training sets.

\begin{table}[h!]
%\vspace*{-3mm}
\centering
\begin{adjustbox}{max width=0.45\textwidth}
\begin{tabular}{| l | c c| c c|} 
 \hline
\multirow{2}{*}{\textbf{Dataset}}  & \multicolumn{2}{|c|}{\textbf{vocab size}} & \multicolumn{2}{|c|}{\textbf{\#tokens/sent}}\\
\cline{2-5}
  & \textbf{src} &\textbf{tgt} &\textbf{src} &\textbf{tgt}\\[0.2ex] 
 \hline\hline
Newsela & 41,066& 30,193& 25.94& 15.89\\  
 WikiSmall & 113,368& 93,835& 24.26& 20.33\\ 
 WikiLarge & 201,841& 168,962& 25.17& 18.51\\ 
 \hline 
\end{tabular}%
\end{adjustbox}
\\[3mm]
\caption{Statistics for the training sets:  the vocabulary size (vocab size), and the average number of tokens per sentence (\#tokens/sent) of the source (src) and target (tgt) language.}
\label{tbl1}
\vspace*{-3mm}
\end{table}

\subsection{Models and Training Details}
We implemented two attention-based Seq2seq models, namely: (1) \textsc{LstmLstm}: the encoder is implemented by two LSTM layers; (2) \textsc{NseLstm}: the encoder is implemented by NSE. The decoder in both cases is implemented by two LSTM layers. The computations for a single model are run on an NVIDIA Titan-X GPU. For all experiments, our models have 300-dimensional hidden states and 300-dimensional word embeddings. Parameters were initialized from a uniform distribution [-0.1, 0.1). We used the same hyperparameters across all datasets. Word embeddings were initialized either randomly or with Glove vectors \cite{Pennington:14} pre-trained on Common Crawl data (840B tokens), and fine-tuned during training. %To test the effectiveness of pre-trained word embeddings, 
We used a vocabulary size of 20K for Newsela, and 30K for WikiSmall and WikiLarge. Our models were trained with a maximum number of 40 epochs using Adam optimizer \cite{Kingma:15} with step size $\alpha = 0.001$ for \textsc{LstmLstm}, and $0.0003$ for \textsc{NseLstm}, the exponential decay rates $\beta_1=0.9,\beta_2=0.999$. The batch size is set to 32. We used dropout \cite{Srivastava:14} for regularization with a dropout rate of 0.3. %The LSTM hidden/memory states of the decoder were initialized with the last LSTM hidden/memory states of the encoder. 
For beam search, we experimented with beam sizes of 5 and 10. Following \cite{Jean:15}, we replaced each out-of-vocabulary token $\langle unk \rangle$ with the source word $x_k$ with the highest alignment score $\alpha_{ti}$, i.e., $k = \argmax \limits_{i} (\alpha_{ti})$.

 Our models were tuned on the development sets, either with BLEU \cite{Papineni:02} that scores the output by counting $n$-gram matches with the reference, or SARI \cite{Xu:16} that compares the output against both the reference and the input sentence. Both measures are commonly used to automatically evaluate the quality of simplification output. 
We noticed that SARI should be used with caution when tuning neural Seq2seq simplification models. Since SARI depends on the differences between a system's output and the input sentence, large differences may yield very good SARI even though the output is ungrammatical. Thus, when tuning with SARI, we ignored epochs in which the BLEU score of the output is too low, using a threshold $\varsigma$. We set $\varsigma$ to 22 on Newsela, 33 on WikiSmall, and 77 on WikiLarge. %We did not use the Flesch-Kincaid Grade Level index \cite[FKGL,][]{Kincaid:75} since it %only takes into account some characteristics of the sentence such as word length and  sentence length, and 
%does not guarantee the grammaticality and meaning preservation of the output \cite{Wubben:12}.
\subsection{Comparing Systems}
We compared our models, either tuned with BLEU (-\textsc{B}) or SARI (-\textsc{S}), against systems reported in \cite{Zhang:17}, namely \textsc{Dress}, a deep reinforcement learning model, \textsc{Dress-Ls}, a combination of \textsc{Dress} and a lexical simplification model \cite{Zhang:17}, \textsc{Pbmt-R}, a PBMT model with dissimilarity-based re-ranking \cite{Wubben:12}, \textsc{Hybrid}, a hybrid semantic-based model that combines a simplification model and a monolingual MT model \cite{Narayan:14}, and \textsc{Sbmt-Sari}, a SBMT model with simplification-specific components. \cite{Xu:16}. %\textsc{Sbmt-Sari} is trained on PPDB \cite{Ganitkevitch:13}, a huge corpus of 106 million sentence pairs and 2 billion words.
\subsection{Evaluation}
%Since there is currently no standard method of evaluating the quality of simplification output, 
%We evaluated system output in two ways, using automatic evaluation measures and human judgments. 
We measured BLEU, and SARI at corpus-level following \cite{Zhang:17}. %In line with previous work \cite{Woodsend:11,Wubben:12,Narayan:14,Xu:16,Zhang:17}
In addition, we also evaluated system output by eliciting human judgments. Specifically, we randomly selected 40 sentences from each test set, and included human reference simplifications and corresponding simplifications from the systems above\footnote{The outputs of comparison systems are available at https://github.com/XingxingZhang/dress.}. We then asked three volunteers\footnote{two native English speakers and one non-native fluent English speaker} to rate simplifications with respect to \textit{Fluency} (the extent to which the output is grammatical English), \textit{Adequacy} (the extent to which the output has the same meaning as the input sentence), and \textit{Simplicity} (the extent to which the output is simpler than the input sentence) using a five point Likert scale.
\section{Results and Discussions}
\subsection{Automatic Evaluation Measures}
The results of the automatic evaluation are displayed in Table \ref{tbl2}. We first discuss the results on Newsela that contains high-quality simplifications composed by professional editors. In terms of BLEU, all neural models achieved much higher scores than \textsc{Pbmt-R} and \textsc{Hybrid}. \textsc{NseLstm-B} scored highest with a BLEU score of 26.31. With regard to SARI, \textsc{NseLstm-S} scored best among neural models (29.58) and came close to the performance of \textsc{Hybrid} (30.00). This indicates that NSE offers an effective means to better encode complex sentences for sentence simplification.

On WikiSmall, \textsc{Hybrid} -- the current state-of-the-art --  achieved best BLEU (53.94) and SARI (30.46) scores. Among  neural models, \textsc{NseLstm-B} yielded the highest BLEU score (53.42), while \textsc{NseLstm-S} performed best on SARI (29.75). On WikiLarge\footnote{Here, BLEU scores are much higher compared to Newsela and WikiSmall since there are 8 reference simplifications for each input sentence in the test set.}, again, \textsc{NseLstm-B} had the highest BLEU score of 92.02. \textsc{Sbmt-Sari} -- that was trained on a huge corpus of 106M sentence pairs and 2B words -- scored highest on SARI with 39.96, followed by \textsc{Dress-Ls} (37.27), \textsc{Dress} (37.08), and \textsc{NseLstm-S} (36.88). 

\begin{table}[t]
\centering
\begin{adjustbox}{max width=0.48\textwidth}
%\resizebox{\columnwidth}{!}{%
\begin{tabular}{| l | c c | c c | c c |} 
 \hline
\multirow{2}{*}{\textbf{Model}}  & \multicolumn{2}{|c|}{\textbf{Newsela}} & \multicolumn{2}{|c|}{\textbf{WikiSmall}} & \multicolumn{2}{|c|}{\textbf{WikiLarge}} \\[0.2ex] 
 \cline{2-7}
& \textbf{BLEU} & \textbf{SARI}  & \textbf{BLEU} & \textbf{SARI}  & \textbf{BLEU} & \textbf{SARI}  \\
 \hline\hline
 \textsc{Pbmt-R} & 18.19 & 15.77 & 46.31 & 15.97 &81.11 & 38.56 \\
 \textsc{Hybrid} & 14.46 & \textbf{30.00} & \textbf{53.94} & \textbf{30.46} & 48.97& 31.40  \\
 \textsc{Sbmt-Sari} & \multicolumn{2}{|c|}{{NA}} & \multicolumn{2}{|c|}{{NA}} & 73.08 & \textbf{39.96} \\
 \textsc{Dress} & 23.21 & 27.37  & 34.53 & 27.48 & 77.18 & 37.08 \\
 \textsc{Dress-Ls} & 24.30& 26.63 & 36.32 & 27.24 &  80.12  & 37.27 \\
  \hline\hline
  \textsc{LstmLstm-B} & 24.38 & 27.66  & 50.53 & 17.67 & 88.81 & 34.22 \\
 \textsc{NseLstm-B} & \textbf{26.31} & 27.42  & 53.42 &17.47 & \textbf{92.02} & 33.43 \\
   \hline\hline
  \textsc{LstmLstm-S} & 23.50 &28.67 & 31.32 & 28.04 & 
 81.95 & 35.45 
\\
 \textsc{NseLstm-S} & 22.62 &29.58 & 29.72 & 29.75  & 80.43 & 36.88 
\\
 \hline
\end{tabular}%
% }
\end{adjustbox}
\\[3mm]
\caption{Model performance using automatic evaluation measures (BLEU and SARI).}
\label{tbl2}
\vspace*{-4mm}
\end{table}
\begin{table*}[ht]
\centering
\begin{adjustbox}{max width=0.7\textwidth}
%\resizebox{\columnwidth}{!}{%
\begin{tabular}{| l | c c c c | c c c c | c c c c |} 
 \hline
\multirow{2}{*}{\textbf{Model}} & \multicolumn{4}{|c|}{\textbf{Newsela}}  & \multicolumn{4}{|c|}{\textbf{WikiSmall}} & \multicolumn{4}{|c|}{\textbf{WikiLarge}} \\[0.2ex] 
 \cline{2-13}
& \textbf{F} & \textbf{A}  & \textbf{S} & \textbf{Avg.}  & \textbf{F} & \textbf{A}  & \textbf{S} & \textbf{Avg.}  &
\textbf{F} & \textbf{A}  & \textbf{S} & \textbf{Avg.}  \\
 \hline\hline
 \textsc{Reference} 
&4.58	&2.98	&3.99	&3.85 &4.63	&3.97	&3.59	&4.06 &4.59	&4.43	&2.38	&3.80 \\
  \hline\hline
 \textsc{Pbmt-R} &3.73	&\textbf{3.90}	&1.98	&3.20 &4.07	&4.11	&2.28	&3.49 &4.22	&4.09	&2.31	&3.54 \\
  \textsc{Hybrid} &2.77	&2.56	&2.41	&2.58 &3.21	&3.62	&2.56	&3.13 &2.63	&2.48	&2.26	&2.46 \\
 \textsc{Sbmt-Sari} &\multicolumn{4}{|c|}{{NA}} & \multicolumn{4}{|c|}{{NA}} &3.89	&3.87	&2.54	&3.43 \\
 \textsc{Dress} &3.98	&2.84	&2.93	&3.25 &4.35	&3.33	&3.49	&3.72 &4.56	&3.66	&2.63	&3.62 \\
 \textsc{Dress-Ls} &3.99	&2.90	&2.98	&3.29 &4.43	&3.33	&3.56	&3.77 &4.68	&3.88	&2.63	&3.73 \\
  \hline\hline
  \textsc{LstmLstm-B} &3.95	&2.93	&3.14	&3.34 & 4.42	&3.88	&2.65	&3.65 &\textbf{4.80}	&4.47	&1.89	&3.72 \\
 \textsc{NseLstm-B} &\textbf{4.26}	&3.13	&3.39	&\textbf{3.59} &\textbf{4.74}	&\textbf{4.22}	&2.49	&3.82 &4.73	&\textbf{4.58}	&1.94	&3.75 \\
   \hline\hline
  \textsc{LstmLstm-S} &4.24	&3.03	&\textbf{3.45} &3.57  &4.59	&3.40	&3.42	&3.80 &4.73	&4.23	&2.21	&3.72 \\
 \textsc{NseLstm-S} &3.83	&2.78	&3.01	&3.21 & 4.57	&3.28	&\textbf{3.81}	&\textbf{3.89} &4.65	&3.95	&\textbf{2.90} &\textbf{3.83} \\

 \hline
\end{tabular}%
% }
\end{adjustbox}
\\[3mm]
\caption{Average human ratings (Fluency (F), Adequacy (A),  Simplicity (S), and Average (Avg.)).}
\label{tbl3}
%\vspace*{-5mm}
\end{table*}
\subsection{Human Judgments}
The results of human judgments are displayed in Table \ref{tbl3}. On Newsela, \textsc{NseLstm-B} scored highest on Fluency. \textsc{Pbmt-R} was significantly better than all other systems on Adequacy while \textsc{LstmLstm-S} performed best on Simplicity. \textsc{NseLstm-B} did very well on both Adequacy and Simplicity, and was best in terms of Average. Example model outputs on Newsela are provided in Table \ref{tbl4}. 

\begin{table*}[ht!]
\centering
\begin{adjustbox}{max width=\textwidth}
%\resizebox{0.4\columnwidth}{!}{%
\begin{tabular}{|p{21cm}|}
 \hline
\textsc{Complex}: Another parent , Mike Munson , sits on the bench with a tablet and uses an app to track and analyze the team 's shots . \\
  \hline
\textsc{Reference}: \textbf{Basketball} parent Mike Munson sits on the bench with a tablet , \textbf{like an iPad} .\\
  \hline
  \textsc{Pbmt-R}: Another parent , Mike Munson \textbf{is} on the bench with a tablet and uses an app to track and analyze the team 's shots .\\
  \hline
  \textsc{Hybrid}: another parent , mike munson sits uses an app to track and analyze shots .\\
  \hline
  \textsc{Dress}: Another parent , Mike Munson , sits on the bench with a \textbf{computer} .\\
  \hline
  \textsc{Dress-Ls}: Another parent , Mike Munson , sits on the bench with a \textbf{computer} .\\
  \hline
  \textsc{LstmLstm-B}: \textbf{He starts} on the bench with a tablet and uses an app to track .\\
  \hline
  \textsc{NseLstm-B}: Another parent , Mike Munson , sits on the bench with a tablet and uses an app to track .\\
  \hline
    \textsc{LstmLstm-S}: \textbf{She} sits on the bench with a tablet and uses an app to track and \textbf{study} the team 's shots .\\
  \hline
    \textsc{NseLstm-S}: \textbf{He} sits on the bench with a tablet .\\
  \hline
\hline
\textsc{Complex}: Stowell believes that even documents about Lincoln 's death will give people a better understanding of the man who was assassinated 150 years ago this April . \\
  \hline
\textsc{Reference}: Stowell \textbf{thinks} that even \textbf{information} about Lincoln 's death will \textbf{help} people \textbf{understand him} .\\
  \hline
  \textsc{Pbmt-R}: Stowell \textbf{thinks} that even documents about Lincoln 's death will give people a better understanding of the man who was killed 150 years ago this April .\\
  \hline
  \textsc{Hybrid}: documents \textbf{that} will give people a understanding the man was assassinated 150 years ago .\\
  \hline
  \textsc{Dress}: Stowell \textbf{thinks} that even documents about Lincoln 's death will give people a better understanding of the man .\\
  \hline
  \textsc{Dress-Ls}: Stowell \textbf{thinks} that even documents about Lincoln 's death will give people a better understanding of the man .\\
  \hline
  \textsc{LstmLstm-B}: Stowell believes that \textbf{only} documents about Lincoln 's death will give people a better understanding .\\
  \hline
  \textsc{NseLstm-B}: Stowell believes that \textbf{the discovery} about Lincoln 's death will give people a better understanding of the man .\\
  \hline
    \textsc{LstmLstm-S}: Stowell \textbf{thinks} that even documents about Lincoln 's death will give people a better understanding of the man .\\
  \hline
    \textsc{NseLstm-S}: Stowell \textbf{thinks} that even \textbf{papers} about Lincoln 's death will give people a better understanding of the man .\\
  \hline
\end{tabular}%
%}
\end{adjustbox}
\\[3mm]
\caption{Example model outputs on Newsela. Substitutions are shown in bold.}
\label{tbl4}
%841 810
\vspace*{-2mm}
\end{table*}

On WikiSmall, \textsc{NseLstm-B} performed best on both Fluency and Adequacy. On WikiLarge, \textsc{LstmLstm-B} achieved the highest Fluency score while \textsc{NseLstm-B} received the highest Adequacy score. In terms of Simplicity and Average, \textsc{NseLstm-S} outperformed all other systems on both WikiSmall and WikiLarge.

As shown in Table \ref{tbl3}, neural models often outperformed traditional systems (\textsc{Pbmt-R}, \textsc{Hybrid}, \textsc{Sbmt-Sari}) on Fluency. This is not surprising given the recent success of neural Seq2seq models in language modeling and neural machine translation \cite{Zaremba:14,Jean:15}. On the downside, our manual inspection reveals that neural models learn to perform copying very well in terms of rewrite operations (e.g., copying, deletion, reordering, substitution), often outputting the same or parts of the input sentence.

Finally, as can be seen in Table \ref{tbl3}, \textsc{Reference} scored lower on Adequacy compared to Fluency and Simplicity on Newsela. On Wikipedia-based datasets, \textsc{Reference} obtained high Adequacy scores but much lower Simplicity scores compared to Newsela. This supports the assertion by previous work \cite{Xu:15} that SEW has a large proportion of inadequate simplifications.

\subsection{Correlations}
Table \ref{tbl5} shows the correlations between the scores assigned by humans and the automatic evaluation measures. There is a positive significant correlation between Fluency and Adequacy (0.69), but a negative significant correlation between Adequacy and Simplicity (-0.64). BLEU correlates well with Fluency (0.63) and Adequacy (0.90) while SARI correlates well with Simplicity (0.73). BLEU and SARI show a negative significant correlation (-0.54). The results reflect the challenge of managing the trade-off between Fluency, Adequacy and Simplicity in sentence simplification. %There is a significant correlation between Fluency and Adequacy, but a negative significant correlation between Adequacy and Simplicity. BLEU correlates well with Fluency and Adequacy while SARI correlates well with Simplicity.
%We can see a positive significant correlation between Fluency and Adequacy (0.69), and a negative significant correlation between Adequacy and Simplicity (-0.64). There is a positive significant correlation between BLEU and Fluency (0.63), as well as between BLEU and Adequacy (0.90) while there is a negative significant correlation between BLEU and Simplicity (-0.56). In contrast, the significant correlations between SARI and Fluency (-0.48) and Adequacy (-0.81) are both in the negative direction. There is a positive significant correlation between SARI and Simplicity (0.73). BLEU and SARI show a negative significant correlation (-0.54). The results reflect the challenge of managing the trade-off between Fluency, Adequacy and Simplicity in sentence simplification. 
\begin{table}[h!]
\centering
\begin{adjustbox}{max width=0.45\textwidth}
%\resizebox{\columnwidth}{!}{%
\begin{tabular}{| l | c c c c |} 
 \hline
 & \textbf{Adequacy} & \textbf{Simplicity} & \textbf{BLEU} & \textbf{SARI}\\
  \hline
   \hline
 \textbf{Fluency} & 0.69$^{**}$ & -0.03 & 0.63$^{**}$ & -0.48$^{**}$\\
 \textbf{Adequacy} & & -0.64$^{**}$ & 0.90$^{**}$ & -0.81$^{**}$\\
 \textbf{Simplicity} & & & -0.56$^{**}$ & 0.73$^{**}$\\
 \textbf{BLEU} & & & & -0.54$^{**}$\\
 
  \hline
\end{tabular}%
% }
\end{adjustbox}
\\[3mm]
\caption{Pearson correlation between the scores assigned by humans and the automatic evaluation measures. Scores marked $^{**}$ are significant at $p < 0.01.$}
\label{tbl5}
\vspace*{-3mm}
\end{table}

\section{Conclusions}
In this paper, we explore neural Seq2seq models for sentence simplification. We propose to use an architecture with augmented memory capacities which we believe is suitable for the task, where one is confronted with long and complex sentences. Results of both automatic and human evaluation on different datasets show that our model is capable of significantly reducing the reading difficulty of the input, while performing well in terms of grammaticality and meaning preservation.
%\bibliography{naaclhlt2018}

\section{Acknowledgements}
We would like to thank Emily Druhl, Jesse Lingeman, and the UMass BioNLP team for their help with this work. We also thank Xingxing Zhang, Sergiu Nisioi for valuable discussions. The authors would like to acknowledge the reviewers for their thoughtful comments and suggestions.
\bibliography{naaclhlt2018}
\bibliographystyle{acl_natbib}

\end{document}